\documentclass[runningheads]{llncs}
\usepackage{graphicx}
\usepackage[utf8]{inputenc} 
\usepackage[T1]{fontenc}    
\usepackage{hyperref}       
\usepackage{url}            
\usepackage{booktabs}       
\usepackage{amsfonts}       
\usepackage{nicefrac}       
\usepackage{microtype}      
\usepackage{lipsum}
\usepackage{fancyhdr}       
\usepackage{graphicx}       
\graphicspath{{media/}}     
\usepackage[T1]{fontenc}
\usepackage{array}
\usepackage{float}
\usepackage{graphicx}
\usepackage{nomencl}
\usepackage{multirow}
\usepackage{dirtytalk}
\usepackage{tabto}
\usepackage{comment}
\usepackage{subcaption}
\usepackage{adjustbox}
\usepackage{amsmath}
\usepackage{adjustbox}
\usepackage{makecell}
\usepackage{longtable}
\usepackage{mathtools}
\usepackage{array,ragged2e}
\usepackage{booktabs}
\usepackage{xcolor}
\usepackage{cite}
\usepackage{arydshln}

\begin{document}
\title{SecureBERT: A Domain-Specific Language Model for Cybersecurity}
%

%

\author{Ehsan Aghaei\inst{1} \and
Xi Niu\inst{1} \and
Waseem Shadid \inst{1} \and
Ehab Al-Shaer \inst{2}}

\institute{{University of North Carolina at Charlotte, USA}\\
\email{\{eaghaei, xniu2, waseem\}@uncc.edu}\\
\and
Carnegie Mellon University, USA\\
\email{ehab@cmu.edu}}

\authorrunning{SecureBERT}
\maketitle            
\begin{abstract}
Natural Language Processing (NLP) has recently gained wide attention in cybersecurity, particularly in Cyber Threat Intelligence (CTI) and cyber automation. 
Increased connection and automation have revolutionized the world's economic and cultural infrastructures, while they have introduced risks in terms of cyber attacks. 
CTI is information that helps cybersecurity analysts make intelligent security decisions, that is often delivered in the form of natural language text, which must be transformed to machine readable format through an automated procedure before it can be used for automated security measures.

This paper proposes SecureBERT, a cybersecurity language model capable of capturing text connotations in cybersecurity text (e.g., CTI) and therefore successful in automation for many critical cybersecurity tasks that would otherwise rely on human expertise and time-consuming manual efforts. SecureBERT has been trained using a large corpus of cybersecurity text.To make SecureBERT effective not just in retaining general English understanding, but also when applied to text with cybersecurity implications, we developed a customized tokenizer as well as a method to alter pre-trained weights. The SecureBERT is evaluated using the standard Masked Language Model (MLM) test as well as two additional standard NLP tasks. Our evaluation studies show that SecureBERT\footnote{\url{https://github.com/ehsanaghaei/SecureBERT}} outperforms existing similar models, confirming its capability for solving crucial NLP tasks in cybersecurity.
\end{abstract}
\section{Introduction}





The adoption of security automation technologies has grown year after year. Cyber security industry is saturated with solutions that protect users from malicious sources, safeguard mission-critical servers, and protect personal information, healthcare data, intellectual property, and sensitive financial data. Enterprises invest in technology to handle such security solutions, typically aggregating a large amount of data into a single system to facilitate in organizing and retrieving key information in order to better identify where they face risk or where specific traffic originates or terminates.
Recently, as social networks and ubiquitous computing have grown in popularity, the overall volume of digital text content has increased. This textual contents span a range of domains, from a simple tweet or news blog article to more sensitive information such as medical records or financial transactions. In cybersecurity context, security analysts analyze relevant data to detect cyber threat-related information, such as vulnerabilities, in order to monitor, prevent, and control potential risks.
For example, cybersecurity agencies such as MITRE, NIST, CERT, and NVD invest millions of dollars in human expertise to analyze, categorize, prioritize, publish, and fix disclosed vulnerabilities annually.
As the number of products grows, and therefore the number of vulnerabilities increases,  it is critical to utilize an automated system capable of identifying vulnerabilities and quickly delivering an effective defense measure.

By enabling machines to swiftly build or synthesize human language, natural language processing (NLP) has been widely employed to automate text analytic operations in a variety of domains including cybersecurity.
Language models, as the core component of modern text analytic technologies, play critical role in NLP applications by enabling computers to interpret qualitative input and transform it into quantitative representations. There are several well-known and well-performing language models, such as ELMO~\cite{peters2018deep}, GPT~\cite{radford2018improving}, and BERT~\cite{devlin2018bert}, trained on general English corpora and used for a variety of NLP tasks such as machine translation, named entity recognition, text classification, and semantic analysis. 
There is continuous discussion in the research community over whether it is beneficial to employ these off-the-shelf models as a baseline, and then fine-tune them through domain-specific tasks. The assumption is that the fine-tuned models will retain the basic linguistic knowledge in general English and meanwhile develop "advanced" knowledge in the domain while fine tuning~\cite{beltagy2019scibert}.

However, certain domains, such as cybersecurity, are indeed highly sensitive, dealing with processing of critical data and any error in this procedure may expose the entire infrastructure to the cyber threats, and therefore, automated processing of cybersecurity text requires a robust and reliable framework.  Cybersecurity terms are either uncommon in general English (such as \textit{ransomware, API, OAuth, exfilterate}, and \textit{keylogger}) or have multiple meanings (homographs) in different domains (e.g., \textit{honeypot, patch, handshake}, and \textit{virus}).
This existing gap in language structure and semantic contexts complicates text processing and demonstrates the standard English language model may be incapable of accommodating the vocabulary of cybersecurity texts, leading to a restricted or limited comprehension of cybersecurity implications.

In this study, we address this critical cybersecurity problem by introducing a new language model called SecureBERT by employing the state-of-the-art NLP architecture called BERT~\cite{devlin2018bert}, which is capable of processing texts with cybersecurity implications effectively.
SecureBERT is generic enough to be applied in a variety of cybersecurity tasks, such as phishing detection \cite{dalton2020active}, code and malware analysis \cite{sajid2020dodgetron}, intrusion detection \cite{aghaei2019host}, etc. SecureBERT is a pre-trained cybersecurity language model that have the fundamental understanding of both the word-level and sentence-level semantics, which is an essential building block for any cybersecurity report. 
In this context, we collected and processed a large corpus of 1.1 billion words (1.6 million in vocabulary size) from a variety of cybersecurity text resources, including news, reports and textbooks, articles, research papers, and videos. On top of the pre-trained tokenizer, we developed a customized tokenization method that preserves standard English vocabulary as much as possible while effectively accommodating new tokens with cybersecurity implication. Additionally, we utilized a practical way to optimize the retraining procedure by introducing random noise to the pre-trained weights.
We rigorously evaluated the performance of our proposed model through three different tasks such as  standard Masked Language Model (MLM), sentiment analysis, and Named Entity Recognition (NER),  to demonstrate SecureBERT's performance in processing both cybersecurity and general English inputs.


\section{Overview of BERT Language Model}
BERT (Bidirectional Encoder Representations from Transformers) \cite{devlin2018bert} is a transformer-based neural network technique for natural language processing pre-training. BERT can train language models based on the entire set of words in a sentence or query (bidirectional training) rather than the traditional way of training on the ordered sequence of words (left-to-right or combined left-to-right and right-to-left). BERT allows the language model to learn word context based on surrounding words rather than just the word that immediately precedes or follows it.

BERT leverages Transformers, an attention mechanism that can learn contextual relations between words and subwords in a sequence. The Transformer includes two separate mechanisms, an encoder that reads the text inputs and a decoder that generates a prediction for the given task. Since BERT\'s goal is to generate a language model, only the encoder mechanism is necessary \cite{vaswani2017attention}. This transformer encoder reads the entire data at the same time instead of reading the text in order. 

Building a BERT model requires two steps: pre-training and fine-tuning. In pre-training stage, the model is trained on unlabeled data against two different pre-training tasks, namely Masked LM (MLM) and Next Sentence Prediction (NSP). MLM typically masks some percentage of the input tokens ($15\%$) at random and then predicts them through a learning procedure. In this case, the final hidden vectors corresponding to the mask tokens are fed into an output softmax over the vocabulary. NSP is mainly designed to understand the relationship between two sentences, which is not directly captured by language modeling. In order to train a model that understands sentence relationships, it trains for a binarized next sentence prediction task that can be trivially generated from any monolingual corpus, in which it takes a pair of sentences as input and in $50\%$ of the times in replaces the second sentence with a random one from the corpus. 
To perform fine-tuning, the BERT model is launched with pre-trained parameters and then all parameters are fine-tuned using labeled data from downstream tasks. BERT model has a unified architecture across different tasks, and there is a minor difference between pre-trained and final downstream architecture. 
The pre-trained BERT model used Books Corpus (800M words) and English Wikipedia (2,500M words) and improved the state-of-the-art for eleven NLP tasks such as getting a GLUE \cite{wang2018glue} score of $80.4\%$, which is $7.6\%$ of definite improvement from the previous best results, and achieving $93.2\%$ accuracy on Stanford Question Answering Dataset (SQuAD) \cite{rajpurkar2016squad}.

A derivative of BERT, which is claimed to be a robustly optimized version of BERT with certain modifications in the tokenizer and the network architecture, and ignored NSP task during training, is called RoBERTa \cite{liu2019roberta}. RoBERTa  extends BERT's MLM, where it intentionally learns to detect the hidden text part inside otherwise unannotated language samples. With considerably bigger mini-batches and learning rates, RoBERTa changes important hyperparameters in BERT training, enabling it to noticeably improve on the MLM and accordingly the overall performance in all standard fine-tuning tasks. 
As a result of the enhanced performance and demonstrated efficacy, we develop SecureBERT on top of RoBERTa.

\section{Data Collection}
We collected a large number ($98,411$) of online cybersecurity-related text data including books, blogs, news, security reports, videos (subtitles), journals and conferences, white papers, tutorials, and survey papers, using our web crawler tool\footnote{Sample data: \url{dropbox.com/sh/jg45zvfl7iek12i/AAB7bFghED9GmkO5YxpPLIuma?dl=0}}. We created a corpus of 1.1 billion words splitting it to 2.2 million documents each with average size of 512 words using the Spacy \footnote{\url{https://spacy.io/usage}} text analytic tool. Table \ref{tab: text_resources} shows the resources and the distribution of our collected dataset for pre-training the SecureBERT.

\begin{table}[!ht]
\setlength\extrarowheight{2pt}
\centering
 \begin{tabular}{|l|c|}
 \hline
 \textbf{Type} & No. \textbf{Documents}\\
 \hline
 \hline
 Articles & 8,955\\
 Books & 180\\
 Survey Papers & 515\\
 Blogs/News & 85,953\\
 Wikipedia (cybersecurity) & 2,156\\
 Security Reports & 518\\
 Videos & 134\\
 \hline
 \textbf{Total} & \textbf{98,411}\\
 \hline
 \end{tabular}
 \quad
 \par\bigskip
 
 \begin{tabular}{|l|c|} 
 \hline
 \textbf{Vocabulary size} & 1,674,434  words \\
 \hline
 \textbf{Corpus size} & 1,072,798,637 words \\
 \hline
 \textbf{Document size} & 2,174,621 documents (paragraphs) \\
 \hline
 \end{tabular}
     \caption{\label{tab: text_resources} The details of collected cybersecurity corpora for training the SecureBERT.}
\end{table}

This corpora contains various forms of cybersecurity texts, from basic information, news, Wikipedia, and tutorials, to more advanced texts such as CTI, research articles, and threat reports. When aggregated, this collection offers a wealth of domain-specific connotations and implications that is quite useful for training a cybersecurity language model. Table \ref{tab: text_resources2} lists the web resources from which we obtained our corpus.
\begin{table}[!ht]
\centering
\setlength\extrarowheight{3pt}
 \begin{tabular}{|l|} 
  \hline
 \multicolumn{1}{|c|}{\textbf{Websites}} \\
\hline
    Trendmicro, NakedSecurity, NIST, GovernmentCIO Media, CShub, Threatpost,\\
    Techopedia, Portswigger, Security Magazine, Sophos, Reddit, FireEye, SANS,\\
    Drizgroup, NETSCOUT, Imperva, DANIEL MIESSLER, Symantec, Kaspersky,\\ 
    PacketStorm, Microsoft, RedHat, Tripwire, Krebs on Security, SecurityFocus, \\
    CSO Online, InfoSec Institute, Enisa, MITRE\\
    \hline
    \hline
    \multicolumn{1}{|c|}{\textbf{Security Reports and Whitepapers}} \\
    \hline
    APT Notes, VNote, CERT, Cisco Security Reports , Symantec Security Reports\\
    \hline\hline
\multicolumn{1}{|c|}{\textbf{Books, Articles, and Surveys}} \\
\multicolumn{1}{|c|}{\textit{Tags: cybersecurity, vulnerability, cyber attack, hack}} \\
\hline
ACM CCS: 2014-2020 , IEEE NDSS (2016-2020), IEEE Oakland (1980-2020)\\
IEEE Security and Privacy (1980-2020), Arxiv , Cybersecurity and Hacking books \\
\hline  \hline
\multicolumn{1}{|c|}{\textbf{Videos (YouTube)}} \\
\hline
Cybersecurity courses, tutorial, and conference presentations\\
\hline
\end{tabular}
     \caption{\label{tab: text_resources2} The resources collected for cybersecurity textual data.}
\end{table}


\section{Methodology}
We present two approaches in this section for refining and training our domain-specific language model. We begin by describing a strategy for developing a customized tokenizer on top of the pre-trained generic English tokenizer, followed by a practical approach for biasing the training weights in order to improve weight adjustment and therefore a more efficient learning process.

\subsection{\textbf{Customized Tokenizer}}\label{sec: Customized Tokenizer}
A word-based tokenizer primarily extracts each word as a unit of analysis, called a token. It assigns each token a unique index, then uses those indices to encode any given sequence of tokens. Pre-trained BERT models mainly return the weight of each word according to these indices. Therefore, in order to fully utilize a pre-trained model to train a specialized model, the common token indices must match, either using the indices of the original or the new customized tokenizer.

For building the tokenizer, we employ a byte pair encoding (BPE) \cite{shibata1999byte} method to build a vocabulary of words and subwords from the cybersecurity corpora, as it is proven to have better performance versus word-based tokenizer. Character based encoding used in BPE allows for the learning of a small subword vocabulary that can encode any input text without introducing any "unknown" tokens \cite{radford2019language}. Our objective is to create a vocabulary that retains the tokens already provided in RoBERTa's tokenizer while also incorporating additional unique cybersecurity-related tokens. In this context, we extract $50,265$ tokens from the cybersecurity corpora to generate the initial token vocabulary $\Psi_{Sec}$. We intentionally make the size of $\Psi_{Sec}$ the same with that of the RoBERTa's token vocabulary $\Psi_{RoBERTa}$ as we intended to imitate original RoBERTa's design.

If $\Psi_{Sec}$ represents the vocabulary set of SecureBERT, and $\Psi_{RoBERTa}$ denotes the vocabulary set of original RoBERTa, both with size of $50,265$, $\Psi_{Sec}$ shares $32,592$ mutual tokens with $\Psi_{RoBERTa}$ leaving $17,673$ tokens contribute uniquely to cybersecurity corpus, such as \textit{firewall, breach, crack, ransomware, malware, phishing, mysql, kaspersky, obfuscated}, and \textit{vulnerability}, where RoBERTa's tokenizer analyzes those using byte pairs:
$$V_{mutual} =\Psi_{Sec}\cap \Psi_{RoBERTa}\rightarrow 32,592 \texttt{ tokens}$$
$$V_{distinct} =\Psi_{Sec}- \Psi_{RoBERTa}\rightarrow 17,673 \texttt{ tokens}$$

Studies~\cite{wang2020neural} shows utilizing complete words (not subwords) for those are common in specific domain, can enhance the performance during training since alignments may be more challenging to understand during model training, as target tokens often require attention from multiple source tokens.
Hence, we choose all mutual terms and assign their original indices, while the remainder new tokens are assigned random indices with no conflict, where the original indices refers to the indices in RoBERTa's tokenizer, to build our tokenizer. 
Ultimately, we develop a customized tokenizer with a vocabulary size similar to that of the original model, which includes tokens commonly seen in cybersecurity corpora in addition to cross-domain tokens. Our tokenizer encodes mutual tokens $V_{mutual}$ as original model, ensuring that the model returns the appropriate pre-trained weights, while for new terms $V_{distinct}$ the indices and accordingly the weights would be random.

\subsection{\textbf{Weight Adjustments}}
The RoBERTa model already stores the weights for all the existing tokens in its general English vocabulary. Many tokens such as \textit{email, internet, computer}, and \textit{phone} in general English convey similar meanings as in the cybersecurity domain. On the other hand, some other homographs such as \textit{adversary, virus, worm, exploit}, and \textit{crack} carry different meanings in different domains. Using the weights from RoBERTa as initial weights for all the tokens, and then re-training against the cybersecurity corpus to update those initial weights will in fact not updating much leading to overfitting condition in training on such tokens because the size of the training data for RoBERTa (16 GB) is 25 times larger than that for SecureBERT. 
When a neural network is trained on a small dataset, it may memorize all training samples, resulting in overfitting and poor performance in evaluation. Due to the unbalance or sparse sampling of points in the high-dimensional input space, small datasets may also pose a more difficult mapping task for neural networks to tackle.

One strategy for smoothing the input space and making it simpler to learn is to add noise to the model during training to increase the robustness of the training process and reduces generalization error.
Referring to previous works on maintaining robust neural networks \cite{zur2009noise, you2019adversarial, liu2018towards}, incorporation of noise to an unstable neural network model with a limited training set can act as a regularizer and help reduce overfitting during the training. It is generally stated that introducing noise to the neural network during training can yield in substantial gains in generalization performance in some cases. Previous research has demonstrated that such noise-based training is analogous to a form of regularization in which an additional term is introduced to the error function \cite{10.1162/neco.1995.7.1.108}.
This noise can be imposed to either input data or between hidden layers of the deep neural networks. When a model is being trained from scratch, typically noise can be added to the hidden layers at each iteration, whereas in continual learning, it can be introduced to input data to generalize the model and reduce error \cite{li2021intelligent, ahn2019uncertainty}.

For training SecureBERT as continual learning process, rather than using the initial weights from RoBERTa directly, we introduce a small "noise" to the weights of the initial model for those mutual tokens, in order to bias these tokens to "be a little away" from the original tokens meanings in order to capture their new connotations in a cybersecurity context, but not "too far away" from standard language since any domain language is still written in English and still carries standard natural language implications. If a token conveys a similar meaning in general English and cybersecurity, the adjusted weight during training will will conceptually tend to converge to the original vector space as the initial model. Otherwise, it will deviate more from the initial model to accommodate its new meaning in cybersecurity. For those new words introduced by the cybersecurity corpus, we use the Xavier weight initialization algorithm \cite{glorot2010understanding} to assign initial weights. 

We instantiated the SecureBERT by utilizing the architecture of pre-trained RoBERTa-base model, which consists of twelve hidden transformer and attention layers, and one input layer. 
We adopted the base version (RoBERTa-base) given the efficiency and usefulness. Smaller models are less expensive to train, and the cybersecurity domain has far less diversity of corpora than general language, implying that a compact model would suffice.
The model's size is not the only factor to consider; usability is another critical factor to consider when evaluating a model's quality. Since large models are difficult to use and expensive to maintain, it is more convenient and practical to use a smaller and portable architecture.

Each input token is represented by an embedding vector with a dimension of $768$ in pre-trained RoBERTa. Our objective is to manipulate these embedding vector representations for each of the $50,265$ tokens in the vocabulary by adding a small symmetric noise. Statistical symmetric noise with a probability density function equal to the normal distribution is known as Gaussian noise. We introduce this noise by applying a random Gaussian function to the weight vectors.
Therefore, for any token $t$, let $\vec{W}_{t}$ be the embedding vector of token $t$as follows:
\begin{equation}
    \label{eq.1}
    \vec{W}_{t}=[w^{t}_1, w^{t}_2,...,w^{t}_{768}]
\end{equation}
where $w^{t}_k$ represents the $k$th element of the embedding vector for token $t$.

Let notation $\mathcal{N}(\mu,\sigma)$ be normal distribution where $\mu$ denotes the mean and $\sigma$ the standard deviation. For each weight vector $\vec{W}_{t}$, the noisy vector $\vec{W}'_{t}$ is defined as follows:
\begin{equation}
    \label{eq.2}    
    \vec{W}'_{t} \leftarrow \vec{W}_{t} \oplus (\vec{W}_{t} \odot \epsilon), \epsilon \sim \mathcal{N}(\mu,\sigma)
\end{equation}
where $\epsilon$ represents the noise value, and $\oplus$ and $\odot$ means element-wise addition and multiplication, respectively.

\begin{figure}
 \center
  \includegraphics[width=9cm]{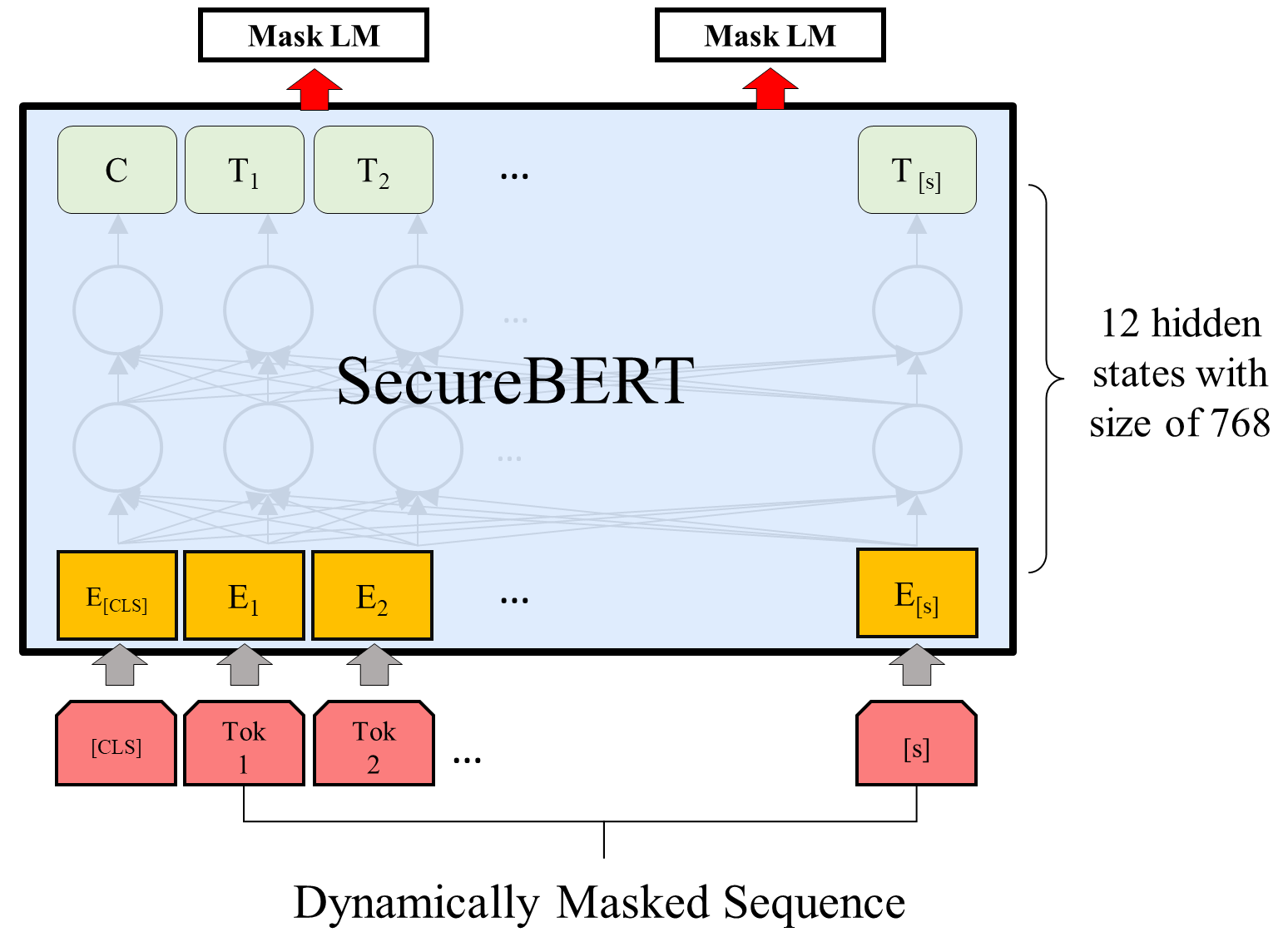}
  \caption{SecureBERT architecture for pre-training against masked words.}
  \label{fig: secbert_model}
\end{figure}

The SecureBERT model is designed to emulate the RoBERTa's architecture, as shown in \ref{fig: secbert_model}. To train SecureBERT for a cybersecurity language model, we use our collected corpora and customized tokenizer. SecureBERT model contains $12$ hidden layers and $12$ attention heads, where the size of each hidden state has the dimension of $768$, and the input embedding dimension is $512$, the same with RoBERTa. In RoBERTa ($768 \times 50265$ elements), the average and variance of the pretrained embedding weights are $-0.0125$ and $0.0173$, respectively.
We picked $mu=0$ and $sigma=0.01$ to generate zero-mean noise value since we want the adjusted weights to be in the same space as the original weights. We replace the original weights in the initial model with the noisy weights calculated using Eq. \ref{eq.2}.

\section{Evaluation}
We trained the model against MLM using dynamic masking using RoBERTa's hyperparameters running for $250,000$ training steps for 100 hours on 8 Tesla V100 GPUs with $Batch\_size=18$, the largest possible mini-batch size for V100 GPUs. We evaluate the model on cybersecurity masked language modeling and other general purpose underlying tasks including sentiment analysis and named entity recognition (NER) to further show the performance and efficiency of SecureBERT in processing the cybersecurity text as well as reasonable effectiveness in general language. 

\subsection{Masked Language Model (MLM)}
In this section, we evaluate the performance of SecureBERT in predicting the masked word in an input sentence, known as the standard Masked Language Model (MLM) task.

\begin{figure}[ht]

  \subfloat[Performance in predicting objects.]{
  \label{fig:mlm_performance1}
	\begin{minipage}[c][1\width]{
	   0.5\textwidth}
	   \centering
	   \includegraphics[width=1\textwidth]{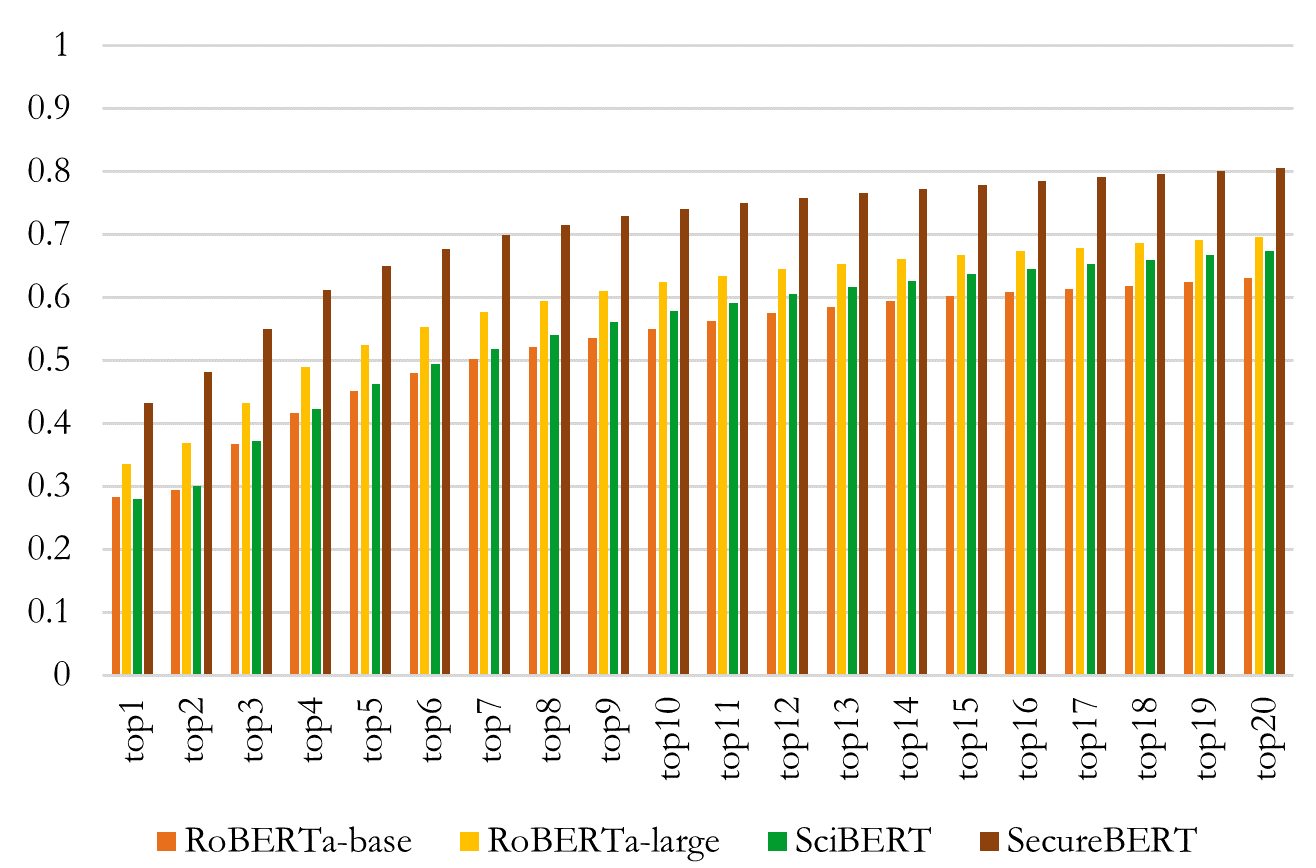}
	\end{minipage}}
 \hfill 	
  \subfloat[Performance in predicting verbs.]{
  \label{fig:mlm_performance2}
	\begin{minipage}[c][1\width]{
	   0.5\textwidth}
	   \centering
	   \includegraphics[width=1\textwidth]{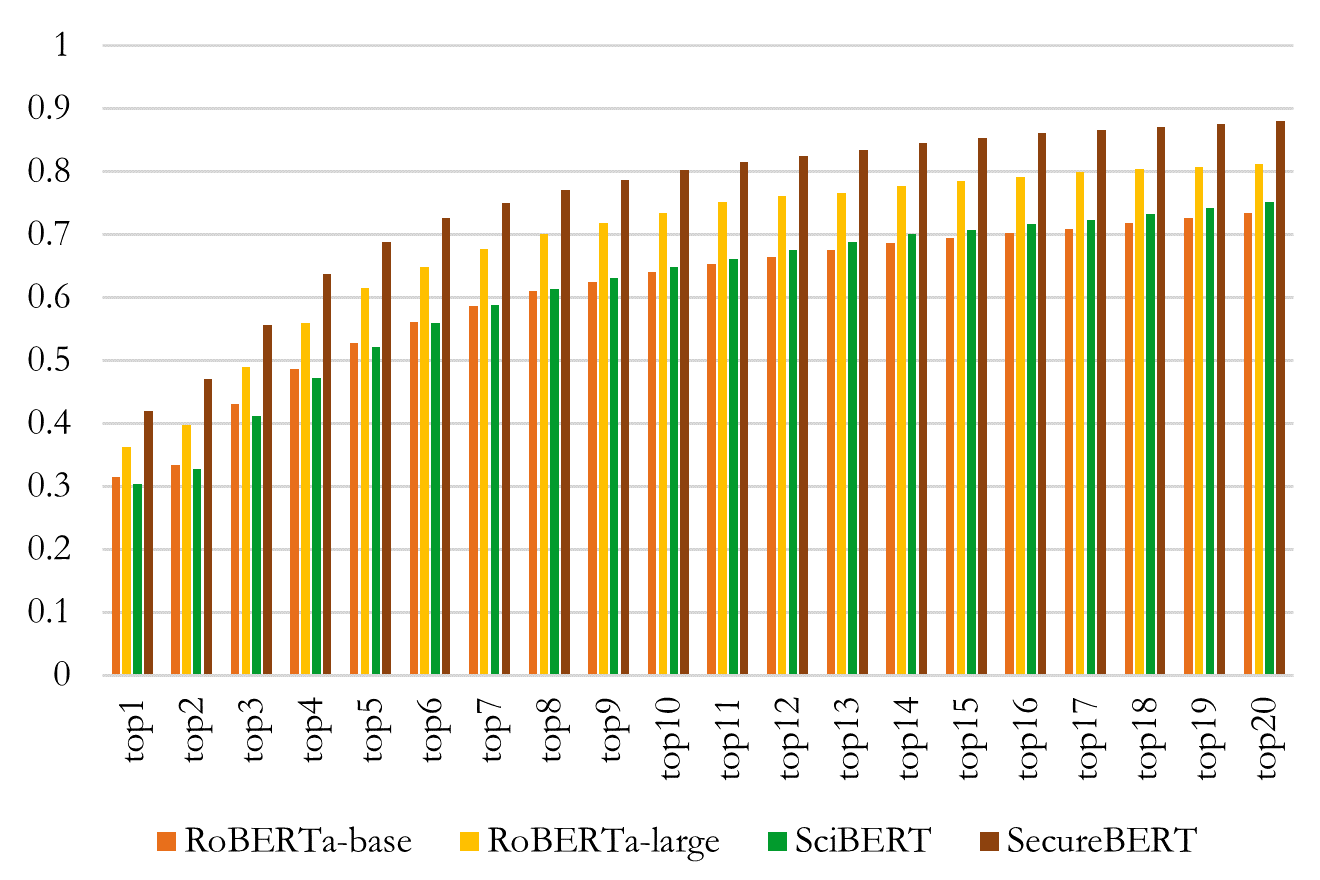}
	\end{minipage}}
 \hfill	
\caption{Cybersecurity masked word prediction evaluation on RoBERTa-base, RoBERTa-large, SciBERT, and SecureBERT.}
\end{figure}

Owing to the unavailability of a testing dataset for the MLM task in the cybersecurity domain, we create one. We extracted sentences manually from a high-quality source of cybersecurity reports - MITRE technique descriptions, which are not included in pre-training dataset. Rather than masking an arbitrary word in a sentence, as in RoBERTa, we masked only the verb or noun in the sentence because a verb denotes an action and a noun denotes an object, both of which are important for understanding the sentence's semantics in a cybersecurity context. Our testing dataset contains $17,341$ records, with $12,721$ records containing a masked noun ($2,213$ unique nouns) and $4,620$ records containing a masked verb ($888$ unique masked verbs in total).
Figure \ref{fig:mlm_performance1} and \ref{fig:mlm_performance2} show the MLM performance for predicting the masked nouns and verbs respectively. Both figures present the prediction hit rate of the masked word in $top N$ model prediction. SecureBERT constantly outperforms RoBERTa-base, RoBERTa-large and SciBERT even though the RoBERTa-large is a considerably large model trained on a massive corpora with $355M$ parameters.

\begin{figure}[!ht]
\centerline{\includegraphics [width=\textwidth]{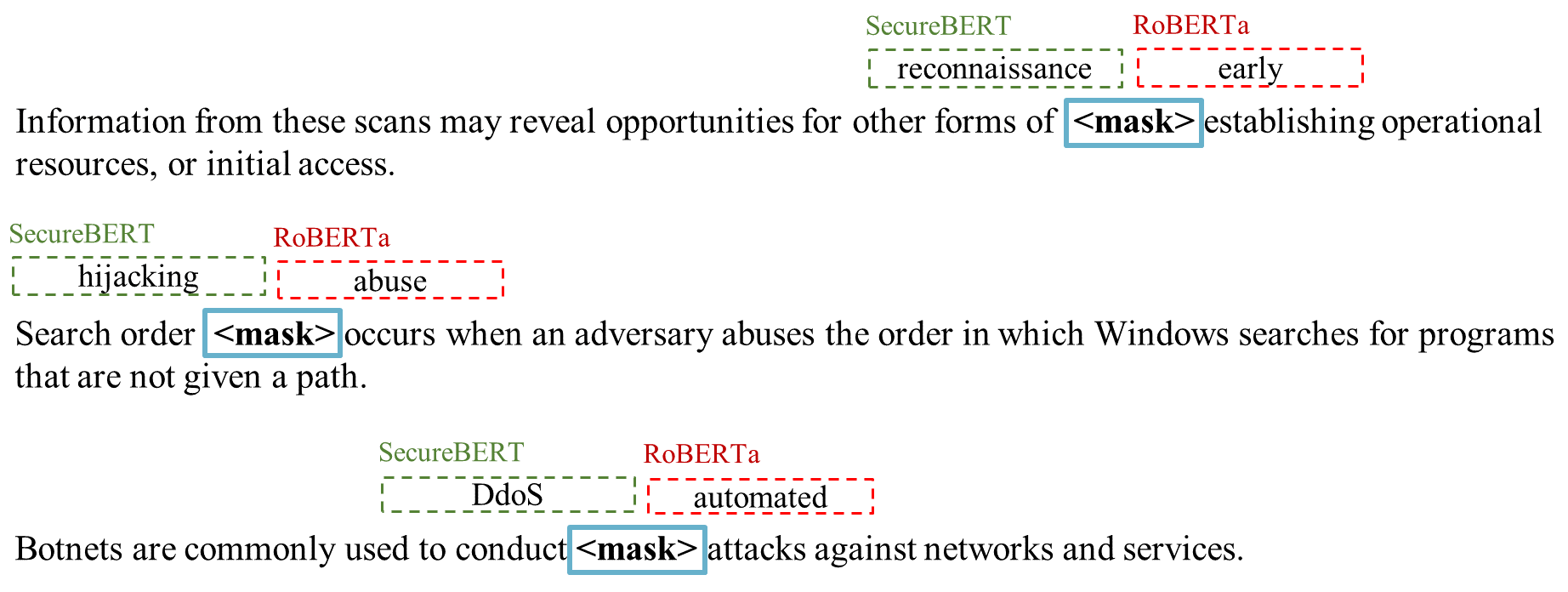}}
\caption {A comparative example of predicting masked token. SecureBERT shows a good understanding of cybersecurity context while other models constantly failed in advanced texts.}
\label{fig: mlm_example}
\end{figure}

Our investigations show that RoBERTa-large (much larger than RoBERTa-base which we used as initial model) is pretty powerful language model in general cybersecurity language. However, when it comes to advance cybersecurity context, it constantly fails to deliver desired output. For example, three cybersecurity sentences are depicted in Fig. \ref{fig: mlm_example}, each with one word masked. Three terms including \textit{reconnaissance, hijacking}, and \textit{DdoS} are commonly used in cybersecurity corpora. SecureBERT is able to understand the context and properly predict these masked words, while RoBERTa's prediction is remarkably different. 
When it comes to cybersecurity tasks including cyber threat intelligence, vulnerability analysis, and threat action extraction \cite{aghaei2019threatzoom,aghaei2020threatzoom}, such knowledge is crucial and utilizing a model with SecureBERT's properties would be highly beneficial. The models do marginally better in predicting verbs than nouns, according to the prediction results. \\
\subsection{Ablation Study}
SecureBERT outperforms existing language models in predicting cybersecurity-related masked tokens in texts, demonstrating its ability to digest and interpret in-domain texts. To enhance its performance and maintain general language understanding, we used specific strategies such as the development of custom tokenizers and weight adjustment. 

SecureBERT employs an effective weight modification by introducing a small noise to the initial weights of the pre-trained model when trained on a smaller corpus than off-the-shelf large models, enabling it to better and more efficiently fit the cybersecurity context, particularly in learning homographs and phrases carrying multiple meanings in different domains.
As a result of the noise, this technique puts the token in a deviated space, allowing the algorithm to adjust embedding weights more effectively.

In Table \ref{tab: MLM_noiseExample}, given a  few simple sentences containing common homographs in cybersecurity context, we provide the masked word prediction of four different models, including SB (SecureBERT), SB* (SecureBERT trained without weight adjustment), RB (RoBERTa-base), and RL (RoBERTa-large). For example, word \textit{Virus} in cybersecurity context refers to a malicious code that spreads between devices to damage, disrupt, or steal data. On the other hand, a \textit{Virus} is also a nanoscopic infectious agent that replicates solely within an organism's live cells. In simple sentence such as "\textit{Virus causes <mask>.}", four models deliver different prediction, each corresponding to associated context. RB and RL return \textit{cancer, infection} and \textit{diarrhea}, that are definitely correct in general (or medical) context, they are wrong in cybersecurity domain though. SB* returns a set of words including \textit{problem, disaster} and \textit{crashes}, which differ from the outcomes of generic models, yet far away from cybersecurity implication. Despite, SB predictions which are \textit{DoS}, \textit{crash}, and \textit{reboot} clearly demonstrate how weight adjustment helps in improved inference of the cybersecurity context by returning the most relevant words for the masked token.

\begin{table}
\footnotesize
\centering
\begin{tabular}{|l | l |} 
  \hline
   \multicolumn{1}{|c|}{\textbf{Masked Sentence}} & \multicolumn{1}{|c|}{\textbf{Model Predictions}}\\
   \hline\hline

   \multirow{4}{*}{Virus causes <mask>.}
        & \textbf{SB}: DoS | crash | reboot \\
        & \textbf{SB*}: problems | disaster | crashes\\
        & \textbf{RB}: cancer | autism | paralysis\\
        & \textbf{RL}: cancer | infection | diarrhea\\
   \hline
   
  \multirow{4}{*}{Honeypot is used in <mask>.}
    & \textbf{SB}: Metasploit | Windows | Squid\\
    & \textbf{SB*}: images | software | cryptography\\
    & \textbf{RB}: cooking | recipes | baking\\
    & \textbf{RL}: cooking | recipes | baking\\
\hline

  \multirow{4}{*}{A worm can <mask> itself to spread.}
    & \textbf{SB}: copy | propagate | program\\
    & \textbf{SB*}: use | alter | modify\\
    & \textbf{RB}: allow | free | help\\
    & \textbf{RL}: clone | use | manipulate\\
\hline

  \multirow{4}{*}{Firewall is used to <mask>.}
    & \textbf{SB}: protect | prevent | detect\\
    & \textbf{SB*}: protect | hide | encrypt\\
    & \textbf{RB}: protect | communicate | defend\\
    & \textbf{RL}: protect | block | monitor\\
\hline

  \multirow{4}{*}{zombie is the other name for a <mask>.}
    & \textbf{SB}: bot | process | trojan\\
    & \textbf{SB*}: worm | computer | program\\
    & \textbf{RB}:  robot | clone | virus\\
    & \textbf{RL}: vampire | virus | person\\
\hline

\end{tabular}\hfil
     \caption{\label{tab: MLM_noiseExample} Shows the masked word prediction results returned by SecureBERT (SB), SecureBERT without weight adjustment (SB*), RoBERTa-base (RB) and RoBERTa-large (RL) in sentences containing homographs}
\end{table}

\begin{figure}[ht]

  \subfloat[Performance in predicting objects.]{
  \label{fig:mlm_performance_compare}
	\begin{minipage}[c][1\width]{
	   0.5\textwidth}
	   \centering
	   \includegraphics[width=1\textwidth]{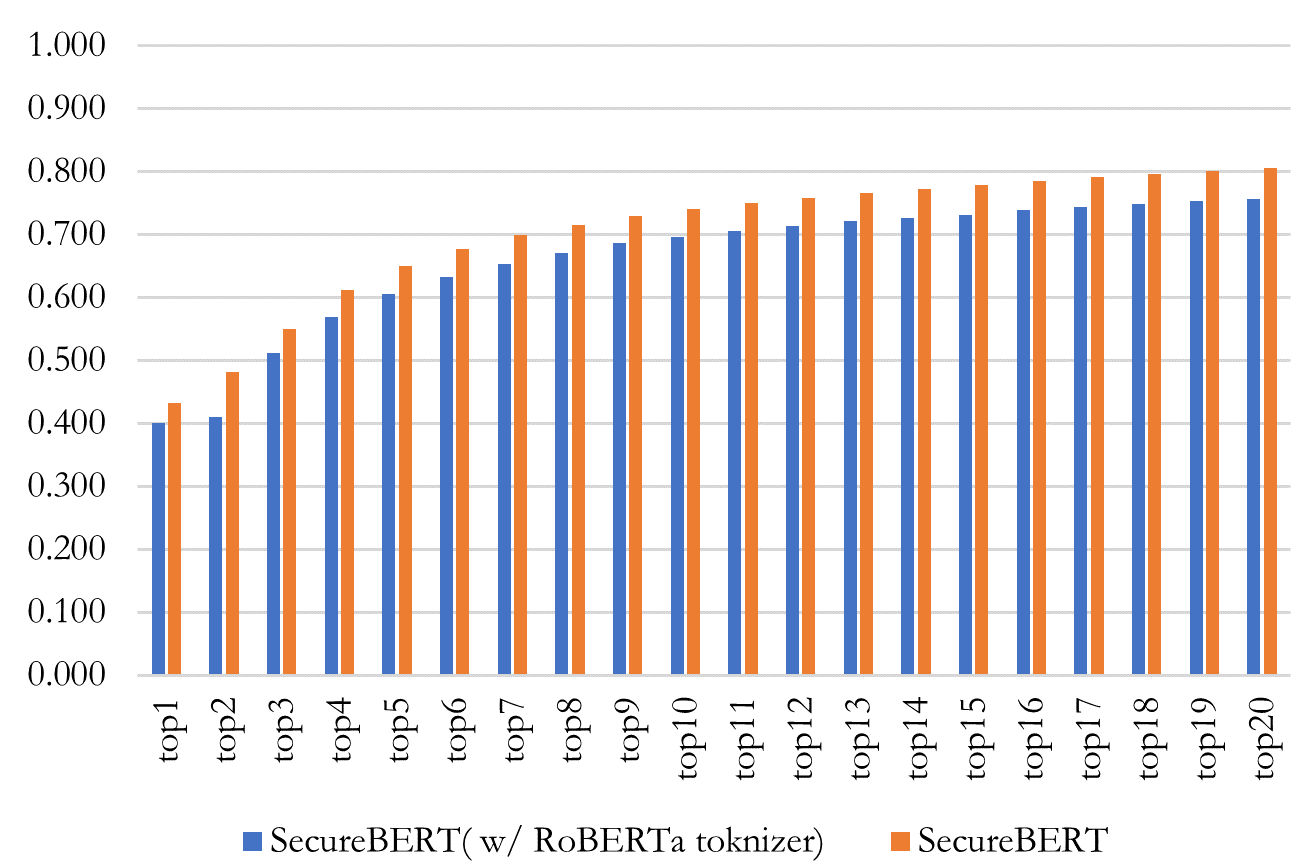}
	\end{minipage}}
 \hfill 	
  \subfloat[Performance in predicting verbs.]{
  \label{fig:mlm_performance2}
	\begin{minipage}[c][1\width]{
	   0.5\textwidth}
	   \centering
	   \includegraphics[width=1\textwidth]{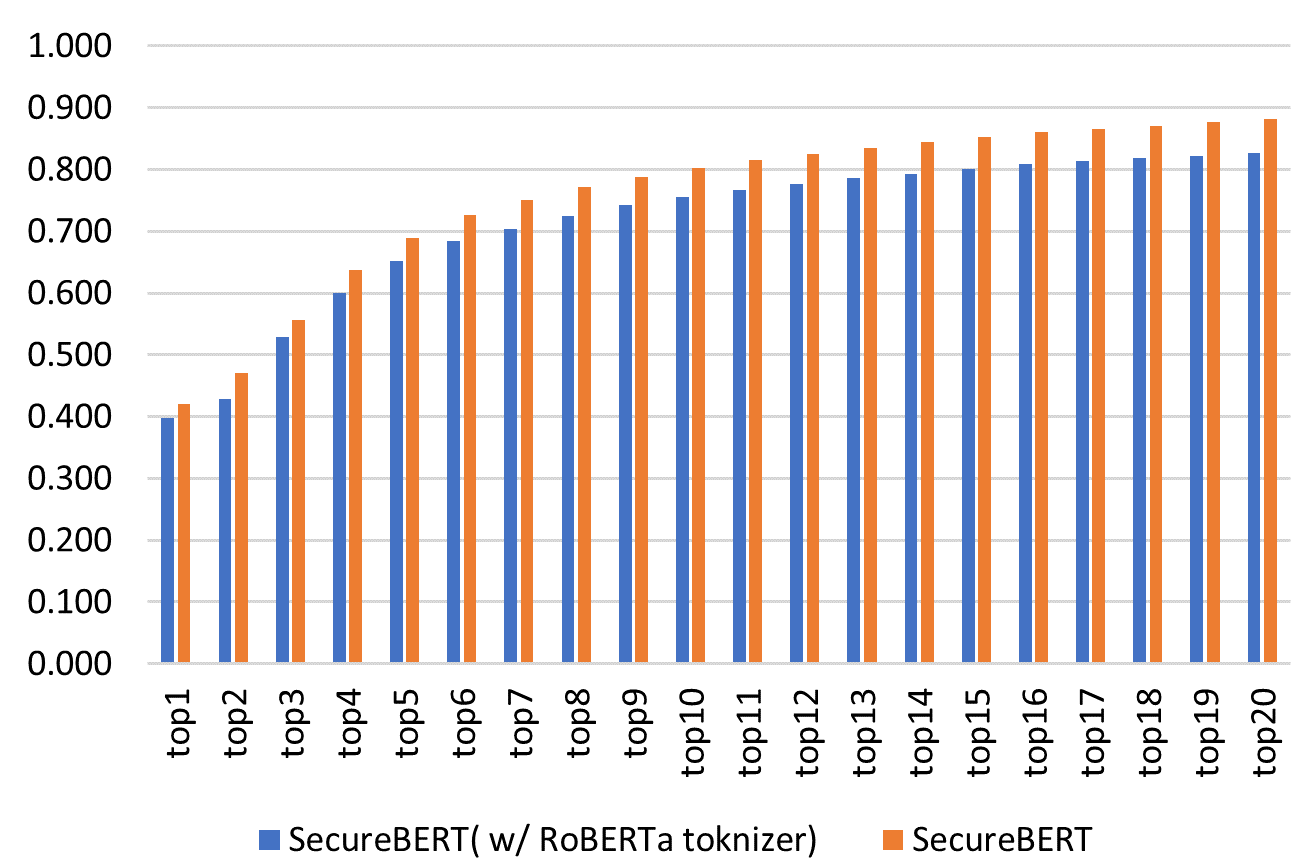}
	\end{minipage}}
 \hfill	
\caption{Demonstrating the impact of the customized tokenizer in masked word prediction performance.}
\end{figure}

Customized tokenizer, on the other hand, also plays an important role in enhancing the performance of SecureBERT in MLM task, by indexing more cybersecurity related tokens (specially complete words as mentioned in Section \ref{sec: Customized Tokenizer}). To further show the impact of SecureBERT tokenizer in returning correct mask word prediction, we train SecureBERT with original RoBERTa's tokenizer without any customization (but with weight adjustment). As depicted in Fig. \ref{fig:mlm_performance_compare} and Fig. \ref{fig:mlm_performance2}, when compared to the pre-trained tokenizer, SecureBERT's tokenizer clearly has a higher hit rate, which highlights the significance of creating a domain-specific tokenizer for any domain-specific language model.

\subsection{Fine-tuning Tasks}
To further proof the performance of SecureBERT in handling the general NLP tasks, we conduct two training experiments including sentiment analysis  as well as named entity recognition (NER).

\subsection*{Task1: Sentiment Analysis}
In the first task, we intend to evaluate the SecureBERT in comprehending general English language in form of sentiment analysis. Thus, we use publicly available Rotten Tomatoes dataset~\footnote{https://www.kaggle.com/c/movie-review-sentiment-analysis-kernels-only} that contains corpus of movie reviews used for sentiment analysis. Socher et al. \cite{socher2013recursive} used Amazon's Mechanical Turk to create fine-grained labels for all parsed phrases in the corpus.
The dataset is comprised of tab-separated files with phrases from the Rotten Tomatoes dataset.  Each Sentence has been parsed into many phrases by the Stanford parser. Each phrase has a "Phrase Id" and each sentence contains a "Sentence Id" while there is no duplicated phrase included in the dataset. Phrases are labeled with five sentiment impressions including  negative, somewhat negative, neutral, somewhat positive, and positive. We build a single layer MLP on top of the four models as  classification layer to classify the phrases to the corresponding  label.
We trained two version of the SecureBERT called raw SecureBERT and modified SecureBERT. The former model is the version of our model in which we utilized customized tokenizer and the weight adjustment method, while the latter is the original RoBERTa model trained as is, using the collected cybersecurity corpora. We trained the model for 1,500 steps with $learning rate = 1e-5$ and $Batch\_size = 32$, to minimize the error of $Cross Entropy$ loss function employing $Adam$ optimizer and $Softmax$ as the activation function in the classification layer. Fig. \ref{fig: sentiment_arch} shows the SecureBERT's architecture for sentiment analysis.

\begin{figure}
 \center
  \includegraphics[width=8cm]{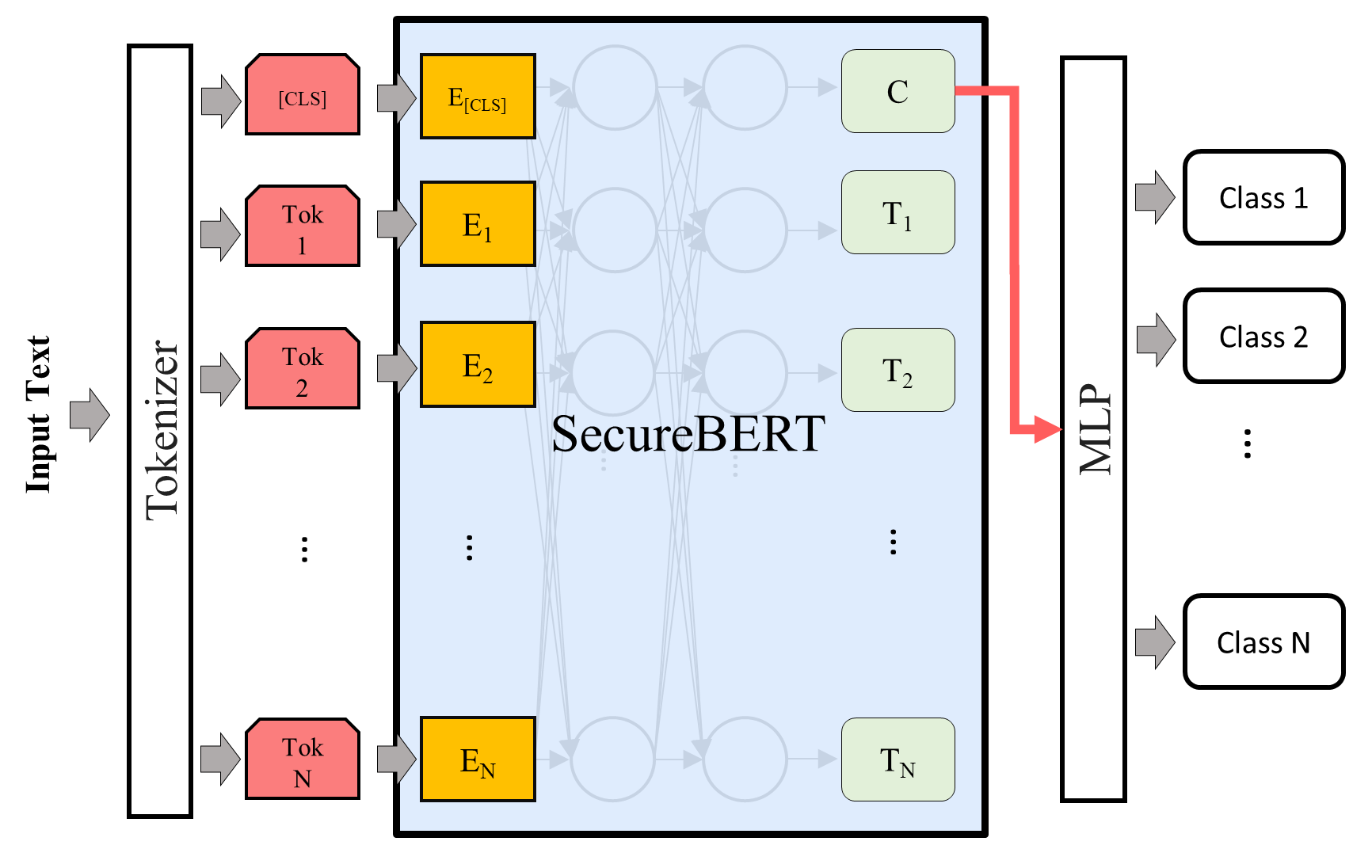}
  \caption{SecureBERT architecture for sentiment analysis downstream task.}
  \label{fig: sentiment_arch}
\end{figure}

In Table \ref{tab: sentiment_analysis1}, we show the performance of both models and compared it with original RoBERTa-base and SciBERT, fine-tuned on Rotten Tomatoes dataset. As illustrated, despite the fact that SciBERT is trained on a broader range of domains (biomedical and computer science), both SecureBERT versions perform quite similarly to SciBERT. In addition, the 2.23\% and 2.02\% difference in accuracy and F1-score with RoBERTa-base demonstrates the effectiveness of SecureBERT in analysing the general English language as well. Furthermore, the modified model perform slightly better than the raw version by 0.34\% accuracy and 0.71\% F1-score improvement. 
\begin{table}[!ht]
\centering
 \begin{tabular}{|l|c|c|c|} 
 \hline
 \textbf{Model Name} & \textbf{Error} & \textbf{Accuracy} & \textbf{F1-Score}\\
 \hline
 \hline
  RoBERTa-base & 0.733 & 69.46 & 69.12\\
  \hline
  SciBERT & 0.768 & 67.76 & 67.08\\
  \hline
  SecureBERT (raw) & 0.788 & 66.89  & 66.39\\
  \hline
  SecureBERT (modified) & 0.771 & 67.23 & 67.10\\
 \hline
 \end{tabular}
     \caption{\label{tab: sentiment_analysis1} Shows the performance of different models on general English sentiment analysis task.}
\end{table}
In the second task, we fine-tune the SecureBERT to conduct cybersecurity-related name entity recognition (NER).
NER is a special task in information extraction that focuses on identifying and classifying named entities referenced in unstructured text into predefined entities such as person names, organizations, places, time expressions, etc.

Since general purpose NER models may not always function well in cybersecurity, we must employ a domain-specific dataset to train an effective model for this particular field. Training a NER model in cybersecurity is a challenging task since there is no publicly available domain-specific data and, even if there is, it is unclear how to establish consensus on which classes should be retrieved from the data. Nevertheless, here we aim to fine-tune the SecureBERT on a relatively small sized dataset that is related to cybersecurity just to show the overall performance and compare it with the existing models.
MalwareTextDB \cite{lim-etal-2017-malwaretextdb} is a dataset containing 39 annotated APT reports with a total of 6,819 sentences. In the NER version of this dataset, the sentences are annotated with four different tags including:\\

\noindent\textbf{Action}: referring to an event, such as "registers", "provides" and "is written".\\

\noindent\textbf{Subject}: referring to the initiator of the Action such as "The dropper" and "This module"\\

\noindent\textbf{Object}: referring  to the recipient of the Action such as "itself ", "remote persistent access" and "The ransom note"; it also refers to word phrases that provide elaboration on the Action such as "a service", "the attacker" and "disk".\\

\noindent\textbf{Modifier}: referring to the tokens that link to other word phrases that provide elaboration on the
Action such as "as" and "to".\\

In each sentence in addition, all the words that are not labeled by any of the mentioned tags as well as pad tokens will be assigned by a dummy label ("O") exclude them in calculating performance metrics.

For Named Entity Recognition, we take the hidden states (the transformer output) of every input token from the last layer from SecureBERT. These tokens are then fed to a fully connected dense layer with \textit{N} units where \textit{N} equals to the total number of defined entities. Since SecureBERT's tokenizer breaks some words into pieces (Bytes), in such cases we just predict the first piece of the word.
\begin{figure}
 \center
  \includegraphics[width=8cm]{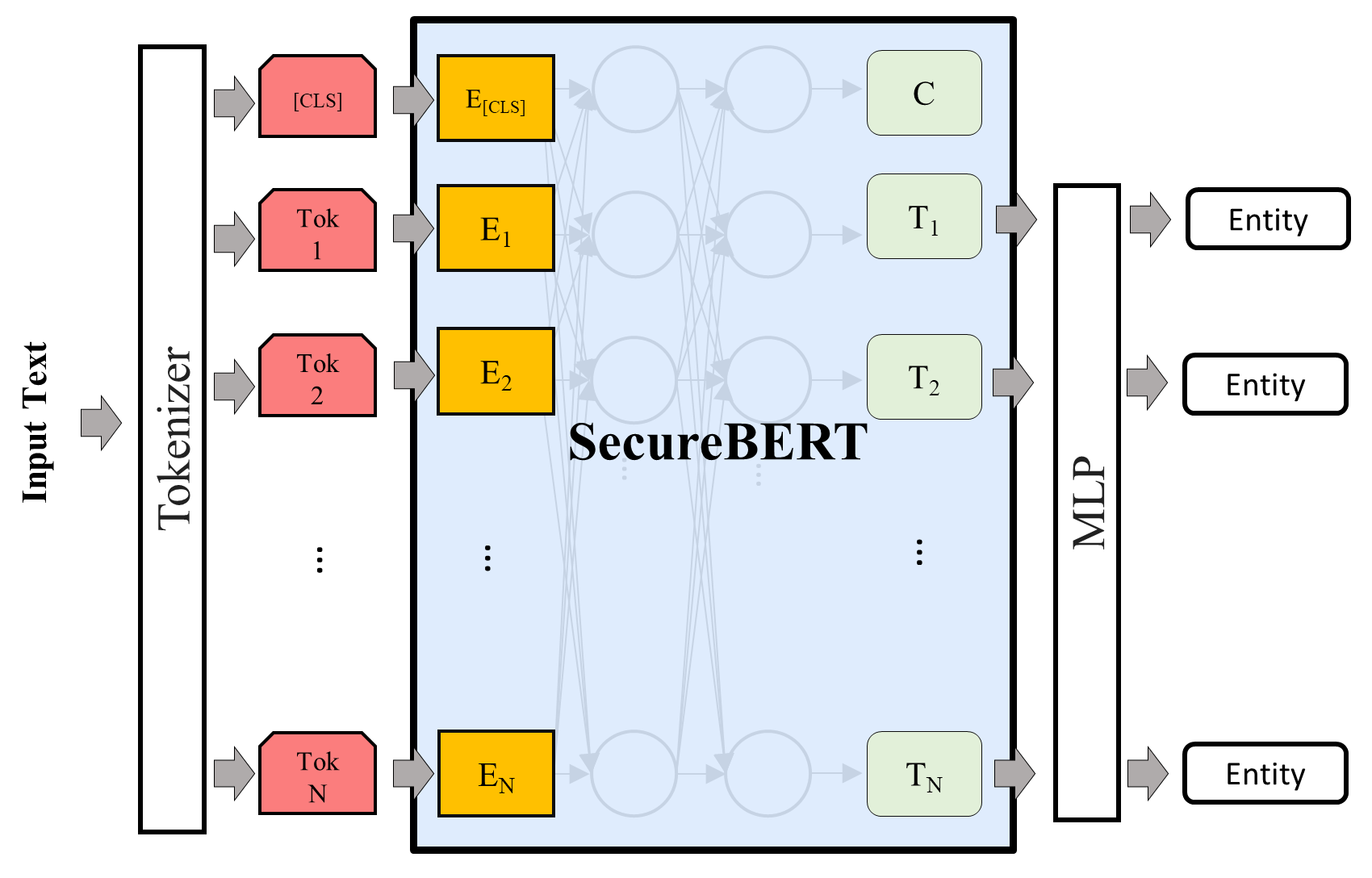}
  \caption{SecureBERT architecture for named entity recognition (NER).}
  \label{fig: sentiment_arch}
\end{figure}

We trained the model in 3 epochs with $learning rate = 2e-5$ and $batch size = 8$, to minimize the error of $Cross Entropy$ loss function using $Adam$ optimizer and $Softmax$ as the activation function in the classification layer.

\begin{table}[!ht]
\centering
 \begin{tabular}{|l|c|c|c|} 
 \hline
 \textbf{Model Name} & \textbf{Precision} & \textbf{Recall} & \textbf{F1-Score}\\
 \hline
 \hline
  RoBERTa-base & 84.92 & 87.53  & 86.20  \\
  \hline
  SciBERT & 83.19 & 85.84 & 84.49\\
  \hline
  SecureBERT (raw) & \textbf{86.08} & 86.81  & 86.44   \\
  \hline
  SecureBERT (modified) & 85.24 & \textbf{88.10}  & \textbf{86.65} \\
 \hline
 \end{tabular}
     \caption{\label{tab: NER_security} Shows the performance of different models trained on MalwareTextDB dataset for NER task.}
\end{table}

Similar to the previous task, Table \ref{tab: NER_security} shows the performance of both SecureBERT's version as well as two other models. As depicted, modified SecureBERT outperforms all other models, despite the fact that MalwareTextDB dataset still contains many sentences with general English meaning and is not an cybersecurity-specific corpora.

\section{Related Works}
Beltagy \textit{et al.} \cite{beltagy2019scibert} unveiled SciBERT following the exact BERT's architecture, a model that improves performance on downstream scientific NLP tasks by exploiting unsupervised pretraining from scratch on a $1.14M$ multi-domain corpus of scientific literature, including $18\%$ computer science and $82\%$ biomedical domain. 

In a similar work on biomedical domain, Gu \textit{et al.}~\cite{lee2020biobert} introduced BioBERT focusing particularly on biomedical domain using BERT architecture and publicly available biomedical datasets. This work also creates a benchmark for biomedical NLP featuring a diverse set of tasks such as named entity recognition, relation extraction, document classification, and question answering. ClinicalBERT \cite{alsentzer2019publicly} is another domain adaptation model based on BERT which is trained on clinical text from the MIMIC-III database. 

Thus far, utilizing language models such as BERT for cybersecurity applications is quite limited.
CyBERT \cite{ameri2021cybert} presents a classifier for cybersecurity feature claims by fine-tuning a pre-trained BERT language model for the purpose of identifying cybersecurity claims from a large pool of sequences in ICS device documents. There are also some other studies working on fine-tuning of BERT in cybersecurity domain. Das \textit{et al.}~\cite{das2021v2w} fine-tunes BERT to hierarchically classify cybersecurity  vulnerabilities to weaknesses. Additionally, there are several studies on fine-tuning BERT for NER tasks such as \cite{chen2021joint}, \cite{zhou2021named} and \cite{gao2021data}. 
Yin \textit{et al.} \cite{yin2020apply} fine-tuned pre-trained BERT against cybersecurity text and developed a classification layer on top of their model, ExBERT, to extract sentence-level semantic features and predict the exploitability of vulnerabilities.
There is also another model called SecBERT \footnote{\url{https://github.com/jackaduma/SecBERT}} published in Github repository which trains BERT on cybersecurity corpus from "APTnotes"\footnote{\url{https://github.com/kbandla/APTnotes}}, "Stucco-Data: Cyber security data sources"\footnote{\url{https://stucco.github.io/data/}}, "CASIE: Extracting Cybersecurity Event Information from Text"\footnote{\url{https://ebiquity.umbc.edu/_file_directory_/papers/943.pdf}}, and "SemEval-2018 Task 8: Semantic Extraction from CybersecUrity REports using Natural Language Processing (SecureNLP)"\footnote{\url{https://ebiquity.umbc.edu/_file_directory_/papers/943.pdf}}. However, at the time of submitting this paper, we could not find any article to learn more about the details and the proof-of-concept to discuss.

\section{Conclusions and Future Works}
This study introduces SecureBERT, a transformer-based language model for processing cybersecurity text language based on RoBERTa.  We presented two practical ways for developing a successful model that can capture contextual relationships and semantic meanings in cybersecurity text by designing a customized tokenization tool on top of RoBERTa's tokenizer and altering the pre-trained weights. SecureBERT is trained to utilize a corpus of 1.1 billion words collected from a range of online cybersecurity resources.
SecureBERT has been evaluated using the standard Masked Language Model (MLM) as well as the named entity recognition (NER) task. The evaluation outcomes demonstrated promising results in grasping cybersecurity language.

\bibliographystyle{splncs04}  
\bibliography{references.bib}  

\begin{thebibliography}{10}
\providecommand{\url}[1]{\texttt{#1}}
\providecommand{\urlprefix}{URL }
\providecommand{\doi}[1]{https://doi.org/#1}

\bibitem{aghaei2019threatzoom}
Aghaei, E., Al-Shaer, E.: Threatzoom: neural network for automated
  vulnerability mitigation. In: Proceedings of the 6th Annual Symposium on Hot
  Topics in the Science of Security. pp.~1--3 (2019)

\bibitem{aghaei2019host}
Aghaei, E., Serpen, G.: Host-based anomaly detection using eigentraces feature
  extraction and one-class classification on system call trace data. Journal of
  Information Assurance and Security (JIAS)  \textbf{14}(4),  106--117 (2019)

\bibitem{aghaei2020threatzoom}
Aghaei, E., Shadid, W., Al-Shaer, E.: Threatzoom: Hierarchical neural network
  for cves to cwes classification. In: International Conference on Security and
  Privacy in Communication Systems. pp. 23--41. Springer (2020)

\bibitem{ahn2019uncertainty}
Ahn, H., Cha, S., Lee, D., Moon, T.: Uncertainty-based continual learning with
  adaptive regularization. Advances in Neural Information Processing Systems
  \textbf{32} (2019)

\bibitem{alsentzer2019publicly}
Alsentzer, E., Murphy, J.R., Boag, W., Weng, W.H., Jin, D., Naumann, T.,
  McDermott, M.: Publicly available clinical bert embeddings. arXiv preprint
  arXiv:1904.03323  (2019)

\bibitem{ameri2021cybert}
Ameri, K., Hempel, M., Sharif, H., Lopez~Jr, J., Perumalla, K.: Cybert:
  Cybersecurity claim classification by fine-tuning the bert language model.
  Journal of Cybersecurity and Privacy  \textbf{1}(4),  615--637 (2021)

\bibitem{beltagy2019scibert}
Beltagy, I., Lo, K., Cohan, A.: Scibert: A pretrained language model for
  scientific text. arXiv preprint arXiv:1903.10676  (2019)

\bibitem{10.1162/neco.1995.7.1.108}
Bishop, C.M.: {Training with Noise is Equivalent to Tikhonov Regularization}.
  Neural Computation  \textbf{7}(1),  108--116 (01 1995).
  \doi{10.1162/neco.1995.7.1.108},
  \url{https://doi.org/10.1162/neco.1995.7.1.108}

\bibitem{chen2021joint}
Chen, Y., Ding, J., Li, D., Chen, Z.: Joint bert model based cybersecurity
  named entity recognition. In: 2021 The 4th International Conference on
  Software Engineering and Information Management. pp. 236--242 (2021)

\bibitem{dalton2020active}
Dalton, A., Aghaei, E., Al-Shaer, E., Bhatia, A., Castillo, E., Cheng, Z.,
  Dhaduvai, S., Duan, Q., Hebenstreit, B., Islam, M.M., et~al.: Active defense
  against social engineering: The case for human language technology. In:
  Proceedings for the First International Workshop on Social Threats in Online
  Conversations: Understanding and Management. pp.~1--8 (2020)

\bibitem{das2021v2w}
Das, S.S., Serra, E., Halappanavar, M., Pothen, A., Al-Shaer, E.: V2w-bert: A
  framework for effective hierarchical multiclass classification of software
  vulnerabilities. In: 2021 IEEE 8th International Conference on Data Science
  and Advanced Analytics (DSAA). pp. 1--12. IEEE (2021)

\bibitem{devlin2018bert}
Devlin, J., Chang, M.W., Lee, K., Toutanova, K.: Bert: Pre-training of deep
  bidirectional transformers for language understanding. arXiv preprint
  arXiv:1810.04805  (2018)

\bibitem{gao2021data}
Gao, C., Zhang, X., Liu, H.: Data and knowledge-driven named entity recognition
  for cyber security. Cybersecurity  \textbf{4}(1),  1--13 (2021)

\bibitem{glorot2010understanding}
Glorot, X., Bengio, Y.: Understanding the difficulty of training deep
  feedforward neural networks. In: Proceedings of the thirteenth international
  conference on artificial intelligence and statistics. pp. 249--256. JMLR
  Workshop and Conference Proceedings (2010)

\bibitem{lee2020biobert}
Lee, J., Yoon, W., Kim, S., Kim, D., Kim, S., So, C.H., Kang, J.: Biobert: a
  pre-trained biomedical language representation model for biomedical text
  mining. Bioinformatics  \textbf{36}(4),  1234--1240 (2020)

\bibitem{li2021intelligent}
Li, X., Yang, Z., Guo, P., Cheng, J.: An intelligent transient stability
  assessment framework with continual learning ability. IEEE Transactions on
  Industrial Informatics  \textbf{17}(12),  8131--8141 (2021)

\bibitem{lim-etal-2017-malwaretextdb}
Lim, S.K., Muis, A.O., Lu, W., Ong, C.H.: {M}alware{T}ext{DB}: A database for
  annotated malware articles. In: Proceedings of the 55th Annual Meeting of the
  Association for Computational Linguistics (Volume 1: Long Papers). pp.
  1557--1567. Association for Computational Linguistics, Vancouver, Canada (Jul
  2017). \doi{10.18653/v1/P17-1143}, \url{https://aclanthology.org/P17-1143}

\bibitem{liu2018towards}
Liu, X., Cheng, M., Zhang, H., Hsieh, C.J.: Towards robust neural networks via
  random self-ensemble. In: Proceedings of the European Conference on Computer
  Vision (ECCV). pp. 369--385 (2018)

\bibitem{liu2019roberta}
Liu, Y., Ott, M., Goyal, N., Du, J., Joshi, M., Chen, D., Levy, O., Lewis, M.,
  Zettlemoyer, L., Stoyanov, V.: Roberta: A robustly optimized bert pretraining
  approach. arXiv preprint arXiv:1907.11692  (2019)

\bibitem{peters2018deep}
Peters, M.E., Neumann, M., Iyyer, M., Gardner, M., Clark, C., Lee, K.,
  Zettlemoyer, L.: Deep contextualized word representations. arXiv preprint
  arXiv:1802.05365  (2018)

\bibitem{radford2018improving}
Radford, A., Narasimhan, K., Salimans, T., Sutskever, I.: Improving language
  understanding by generative pre-training  (2018)

\bibitem{radford2019language}
Radford, A., Wu, J., Child, R., Luan, D., Amodei, D., Sutskever, I., et~al.:
  Language models are unsupervised multitask learners. OpenAI blog
  \textbf{1}(8), ~9 (2019)

\bibitem{rajpurkar2016squad}
Rajpurkar, P., Zhang, J., Lopyrev, K., Liang, P.: Squad: 100,000+ questions for
  machine comprehension of text. arXiv preprint arXiv:1606.05250  (2016)

\bibitem{sajid2020dodgetron}
Sajid, M.S.I., Wei, J., Alam, M.R., Aghaei, E., Al-Shaer, E.: Dodgetron:
  Towards autonomous cyber deception using dynamic hybrid analysis of malware.
  In: 2020 IEEE Conference on Communications and Network Security (CNS).
  pp.~1--9. IEEE (2020)

\bibitem{shibata1999byte}
Shibata, Y., Kida, T., Fukamachi, S., Takeda, M., Shinohara, A., Shinohara, T.,
  Arikawa, S.: Byte pair encoding: A text compression scheme that accelerates
  pattern matching  (1999)

\bibitem{socher2013recursive}
Socher, R., Perelygin, A., Wu, J., Chuang, J., Manning, C.D., Ng, A.Y., Potts,
  C.: Recursive deep models for semantic compositionality over a sentiment
  treebank. In: Proceedings of the 2013 conference on empirical methods in
  natural language processing. pp. 1631--1642 (2013)

\bibitem{vaswani2017attention}
Vaswani, A., Shazeer, N., Parmar, N., Uszkoreit, J., Jones, L., Gomez, A.N.,
  Kaiser, {\L}., Polosukhin, I.: Attention is all you need. In: Advances in
  neural information processing systems. pp. 5998--6008 (2017)

\bibitem{wang2018glue}
Wang, A., Singh, A., Michael, J., Hill, F., Levy, O., Bowman, S.R.: Glue: A
  multi-task benchmark and analysis platform for natural language
  understanding. arXiv preprint arXiv:1804.07461  (2018)

\bibitem{wang2020neural}
Wang, C., Cho, K., Gu, J.: Neural machine translation with byte-level subwords.
  In: Proceedings of the AAAI Conference on Artificial Intelligence. vol.~34,
  pp. 9154--9160 (2020)

\bibitem{yin2020apply}
Yin, J., Tang, M., Cao, J., Wang, H.: Apply transfer learning to cybersecurity:
  Predicting exploitability of vulnerabilities by description. Knowledge-Based
  Systems  \textbf{210},  106529 (2020)

\bibitem{you2019adversarial}
You, Z., Ye, J., Li, K., Xu, Z., Wang, P.: Adversarial noise layer: Regularize
  neural network by adding noise. In: 2019 IEEE International Conference on
  Image Processing (ICIP). pp. 909--913. IEEE (2019)

\bibitem{zhou2021named}
Zhou, S., Liu, J., Zhong, X., Zhao, W.: Named entity recognition using bert
  with whole world masking in cybersecurity domain. In: 2021 IEEE 6th
  International Conference on Big Data Analytics (ICBDA). pp. 316--320. IEEE
  (2021)

\bibitem{zur2009noise}
Zur, R.M., Jiang, Y., Pesce, L.L., Drukker, K.: Noise injection for training
  artificial neural networks: A comparison with weight decay and early
  stopping. Medical physics  \textbf{36}(10),  4810--4818 (2009)

\end{thebibliography}
\end{document}